\documentclass[runningheads]{llncs}
\usepackage{subfigure}
\usepackage{setspace}
\usepackage{geometry} 
\usepackage{epstopdf}
\usepackage[labelsep=period]{caption}  
\usepackage[british]{babel} 
\usepackage[hang]{footmisc}
\usepackage{makecell}
\usepackage{booktabs}
\usepackage[numbers]{natbib}
\usepackage{graphicx}
\usepackage[utf8]{inputenc}

\begin{document}

\title{Can a Large Language Model Assess Urban Design Quality? Evaluating Walkability Metrics Across Expertise Levels}
\date{}


\author{Chenyi Cai\inst{1} \and
Kosuke Kuriyama\inst{1,2} \and
Youlong Gu \inst{1,3}  \and
Filip Biljecki\inst{3} \and
Pieter Herthogs\inst{1}}
\authorrunning{C. Cai et al.}

\institute{Singapore-ETH Centre, Future Cities Lab Global Programme, CREATE campus, 1 Create Way, \#06-01 CREATE Tower, 138602, Singapore. \email{(chenyi.cai, kosuke.kuriyama, youlong.gu, pieter.herthogs)@sec.ethz.ch} \and
 Takenaka Corporation, Project Development Division, 4-1-13 Hommachi, Chuo-Ku, Osaka, 541-0053, Japan. 
 \email{kuriyama.kousuke@takenaka.co.jp} \and
 Department of Architecture, National University of Singapore, 4 Architecture Dr, 117566, Singapore \email{(youlong,filip )@n.nus.edu} }




\maketitle 
\begin{abstract}
Urban street environments are vital to supporting human activity in public spaces. The emergence of big data, such as street view images (SVI) combined with multi-modal large language models (MLLM), is transforming how researchers and practitioners investigate, measure, and evaluate semantic and visual elements of urban environments.  
Considering the low threshold for creating automated evaluative workflows using MLLM, it is crucial to explore both the risks and opportunities associated with these probabilistic models. 
In particular, the extent to which the integration of expert knowledge can influence the performance of MLLM in the evaluation of the quality of urban design has not been fully explored. 
This study set out an initial exploration of how integrating more formal and structured representations of expert urban design knowledge (e.g., formal quantifiers and descriptions from existing methods) into the input prompts of an MLLM (ChatGPT-4) can enhance the model's capability and reliability to evaluate the walkability of built environments using SVIs. 
We collect walkability metrics through the existing literature and categorise them using relevant ontologies. 
Then we select a subset of these metrics, used for assessing the subthemes of pedestrian safety and attractiveness, and develop prompts for MLLMs accordingly. 
We analyse MLLM’s abilities to evaluate SVI walkability subthemes through prompts with multiple levels of clarity and specificity about evaluation criteria. 
Our experiments demonstrate that MLLMs are capable of providing assessments and interpretations based on general knowledge and can support the automation of image-text multimodal evaluations. However, they generally provide more optimistic scores and can make mistakes when interpreting the provided metrics, resulting in incorrect evaluations. By integrating expert knowledge, MLLM's evaluative performance exhibits higher consistency and concentration. Therefore, this paper highlights the importance of formally and effectively integrating domain knowledge into MLLMs for evaluating urban design quality.


\end{abstract}


\section{Introduction}\label{Intro}
 
\sloppy

Urban street environments are vital to supporting human activity in public spaces ~\citep{jacobs2010dark}. Streets are essential connectors within urban networks, 
allowing for seamless movement of pedestrians, cyclists, and vehicles. Well-designed urban environments can have a range of positive impacts, such as encouraging physical activities, improving mood, strengthening urban identity, and promoting public health ~\citep{koohsari2020localwalkabilitycardiovasculardisease,wedyan2025assessingimpactofwalkabilityhealth}. In contrast, poorly designed or unpleasant urban environments can have various negative consequences \citep{de2023spatialcharacterofunpleasantcycling,giles2016cityplanningandpopulationhealth}.

Multi-modal large language models (MLLM) such as GPT-4 (OpenAI), with their ability to analyze textual and visual data, hold the potential for providing evidence-based evaluations and quality improvement suggestions for the built environments. Currently, the large availability of street view imagery (SVI) provides a rich data source, along with the rapid development of computer vision techniques, enabling the computational assessment of the visual quality of the street environment. 
Existing literature on SVI and environmental quality has primarily focused on identifying key correlations between visual street features and travel behaviour, enhancing the accuracy of street quality indicators, and mapping the spatial distribution of environmental attributes within study cities ~\citep{biljecki2021SVIinurbananalytics,liu2024clarityorconfustion}. 
MLLMs hold significant potential for extracting physical environment features and automating evaluative workflows \citep{malekzadeh2025urbanattractivenessChatGPT}. 
Although LLMs demonstrate a certain level of knowledge about global cities, their limitations become evident when they encounter unfamiliar tasks, often producing generic or random outputs \citep{li2024canLLMtellusAboutCities}.
A significant challenge lies in the gap between the generalised training data of MLLMs and the specialised knowledge required for evaluating built environment quality. The risks and advantages of MLLM --- a black box --- in providing evaluation results, and the potential to enhance its performance by integrating expert urban design knowledge, have not been fully explored. 

Formal definitions and representations of expert knowledge enable the automation of environmental quality assessments and the operationalization of urban design. 
Public space quality assessments, such as walkability metrics, utilise determinants and criteria as representations of expert knowledge~\citep{fonseca2022builtenvironmentattributes,ariffin2021systematicreviewofWalkandBE}.
While the key components of walkable urban environments are well-recognised by the research community~\citep{ewing2009measuringthemeasurable,dragovic2023LRofparameter-basedmodelforWB}, translating these principles into interpretable and robust indicators requires formal definitions of domain-specific concepts. 
In walkability evaluation, there are mixed-use of terms and measurements from different domains, scales and data sources. 
First, for instance, land use density is linked to attractiveness in one framework \citep{frank2010developmentWI} and to accessibility in another~\citep{pelclova2014Gender-specificassociations}. Additionally, methods for measuring land use density vary across different studies.
Second, walkability metrics contain both contextual factors (e.g., entropy index of different land uses in an area) and site-specific factors (e.g., presence of streetlights) \citep{fonseca2022builtenvironmentattributes}.
Metrics include elements that are directly measurable (e.g., presence of fixed furniture on streets) and those requiring monitoring (e.g., history of thefts). 
Some elements in these frameworks are actionable for urban design, while others are not \citep{dragovic2023LRofparameter-basedmodelforWB, reisi2019localWI}.
In summary, the variable use of terms and measurements across different frameworks causes challenges for interpretability and comparability between cases and between frameworks. In turn, a lack of formal definitions and categorisations for urban design-related metrics also makes it difficult to generate clear design recommendations to improve environments. 

Exploring the risks and opportunities of MLLM in urban evaluative frameworks through the integration of formal expertise is crucial.  
First, it requires translating multiple characteristics into formalised and methodologically practical indicators. 
Second, integrating expert urban design knowledge (e.g., definitions, categorisations, scoring models) formally and effectively into MLLMs and enhancing their performance in evaluating built environment quality requires more investigations. Hence, the following research questions are raised:
\begin{itemize}
    \item Can MLLMs provide consistent responses when prompted to evaluate street environment quality?
    \item To what extent can expert knowledge (more formal definitions and semantic clarity) and a structured evaluation framework enhance MLLMs' ability to evaluate street environment quality (e.g. walkability)?
\end{itemize}

Our research aims to bridge core urban design variables of street environments with urban design solutions to increase suitability for human activities. We investigate how formal representation of urban design knowledge can improve the MLLM’s consistency to assess walkability. 
In this paper, we sets out the initial explorations and implementations, focusing on: 1) categorising walkability metrics through tangible measurements in the built environment, and 2) investigating how the level of formality in urban design expert knowledge influences MLLM performance, including expert descriptions from the literature review and semantic clarity of walkability metrics.
First, through the existing literature, we collect and categorise walkability metrics based on measurements, criteria, methods, and data sources.
Second, using example metrics related to pedestrian Safety and Attractiveness, we develop prompts, with varying levels of formalization and identification, to compare MLLMs' produced assessments.
Third, along with SVIs from selected locations in Singapore, we examine the performance of MLLMs in evaluating SVI walkability. We apply statistical analysis to the assessment results, focusing on the general score distributions and notable differences between particular metrics.
Finally, the paper provides an example of identifying potential urban design interventions for places requiring improvement, focusing on the metrics that are actionable within the scope of urban design.

\section{Background}\label{sec:bg}

Extensive research explores how urban spatial characteristics influence the suitability of street environment for people and activities. 
Street view images (SVI) enable the study of the physical environment and its interactions with the socioeconomic environment at various scales \citep{zhang2024urbanvisualintelligence} and have been widely used for numerous applications – ranging from analysing vegetation and transportation to health and socio-economic studies \citep{biljecki2021SVIinurbananalytics}. In walkability-related studies, because of SVIs' convenience, field auditing works have been replaced by the desktop-auditing tools \citep{larranaga2019usingbestworst}. 
A wide range of walkability and SVI-related research focuses on uncovering the most significant correlations between visual street features and travel behaviour, as well as mapping the spatial distribution of specific environmental features \citep{larranaga2019usingbestworst,huang2024impactofphysicalfeatures}. 

The advent of MLLMs, such as GPT-4 \citep{achiam2023gpt4}, which combine the textual interaction capabilities of LLMs with image analysis, unlocked new possibilities for applications that demand integrated visual and textual interpretation. 
\citet{liu2023CLIPwalkability} introduced methods using the Multimodal Contrastive Learning Model(CLIP) to assess perceived walkability by analysing both tangible and subjective factors such as safety and attractiveness. Compared with convolutional neural networks (CNN), MLLM has the strength in increasing explainability in AI-driven assessments. 
It can generate interpretations and explanations for walkability assessments, hence it can provide insights into the specific factors that influence the evaluations \citep{blevcic2024enhancingurbanWB}. 
However, it is evident that while LLMs possess a certain level of urban knowledge, they have limitations when faced with unfamiliar tasks, often generating generic or random outputs \citep{li2024canLLMtellusAboutCities}. 
Given the potential of using MLLMs, they are developed based on generalised training data \citep{bender2021dangersofstochasticparrots}. However, evaluating walkability and providing suggestions for improvement requires specialised urban design knowledge. Therefore, whether MLLMs can effectively evaluate walkability and offer professional improvement suggestions and how to integrate expert knowledge remains unexplored.

Various evaluation frameworks and metrics have been developed to represent urban design knowledge, using varied criteria and metrics in different scales.
Focusing on walkability, the metrics from the literature examine local elements (e.g. visible amenities, greenery) and contextual factors (e.g. street networks, neighbourhood land use) \citep{fonseca2022builtenvironmentattributes,dragovic2023LRofparameter-basedmodelforWB}, 
some of which are actionable for urban designers (e.g. park and green zones, aesthetics of buildings), while others are not (e.g. motorised transport speed, weather conditions).
\textit{Greenness} is an example of a criterion name label that is measured in very different ways, such as assessing landscape coverage from satellite images \citep{fan2018walkabilityinurbanlandscape}, calculating the percentage of trees in street view images \citep{huang2024impactofphysicalfeatures}, counting the number of street segments with street trees \citep{lee2020schoolwalkability}, or even defining "green" to include the presence of amenities that seem wholly unrelated to green, such as health services, banks, or auto services \citep{pereira2020relationshipbetweenBEandHealth}.

In many urban design-related walkability evaluation frameworks, there are different sets of criteria. 
\cite{yin2017streetlevelUDforWB} developed the evaluation framework by methodologically interpreting and translating the criteria proposed by \citet{ewing2009measuringthemeasurable}.
\citet{arellana2020WBconsideringperceptionofBE} assess it based on factors such as sidewalk condition, traffic safety, comfort, and attractiveness. Meanwhile, \citet{larranaga2019usingbestworst} employed a different set of criteria, such as connectivity, topography, sidewalk surface, number of police officers, and number of shops.
There is a mix of criteria types. For example, criteria such as safety or attractiveness are dispositions, while criteria such as the number of shops or sidewalk width are directly measurable properties.
Metrics often share the same definition but are labelled differently, for instance, the number of parks was used to measure imageability \citep{yin2017streetlevelUDforWB} in one study and to measure amenity density \citep{fonseca2022builtenvironmentattributes} in another. \cite{grisiute2024ontology} highlighted the need for a structured and shared vocabulary for bike network evaluations to enhance the coherence of evaluation methods within the field of bike network planning --- the same is true for walkability.
The fact that different studies use different evaluation frameworks and models is to be expected, but the observable lack of formal definitions and semantic accuracy. These inconsistencies also hinders the potential implementations using digital technologies (e.g.LLMs).


\begin{tiny}
\begin{table*} [h!]
\noindent
\centering
\caption{Overview of the identified metrics for safety and attractiveness (as used in walkability studies). The vague metrics (Metric-1) exhibit vague definitions in their names and the quantified metrics (Metric-2) have more formal names using explicit quantifiers.}\label{tab1}
\makebox[\textwidth][c]{%
\resizebox{\textwidth}{!}{
\begin{tabular}{ll|ll}
\toprule
 \textbf{\makecell[l]{Vague \\ Safety Metric-1 }} & \textbf{\makecell[l]{Quantified \\ Safety Metric-2}}  & \textbf{\makecell[l]{Vague \\ Attractiveness Metric-1 }}  & \textbf{\makecell[l]{Quantified \\ Attractiveness Metric-2} } \\
\midrule
CrossingAids &  PresenceOfCrossingAids &  GreenArea &  PresenceOfGreenArea  \\
TrafficSignals &  PresenceOfPedestrianSignals &  InstitutionalArea & PresenceOfInstitutionalArea \\
SpeedBumps & PresenceOfTrafficCalmingDevice  & ResidentialArea &  PresenceOfResidentialArea \\
PoliceStations &  PresenceOfPoliceStations &  CommercialArea & PresenceOfCommercialArea\\
CCTV & PresenceOfSecurityCameras & Parks & PresenceOfParks \\
CrossRoads & NumberOfCrossingFacilities & Trees & PresenceOfTrees \\
RiskOfTrafficAccidents & PerceivedRiskOfTrafficAccidents & Attractiveness & PerceivedAttractiveness \\
VehicleSpeed &  MotorisedTransportSpeed & EstheticFeatures & PerceivedNeighborEstheticFeature \\
VehicleFlow & VehicleFlow & CulturalCentres & NumberOfCulturalCentres \\
TrafficSafety & PerceivedTrafficSafety & Retails & AreaOfRetailTradeOrGastronomy \\
TrafficCalmingDevice & NumberOfTrafficCalmingDevices & FixedFurniture & NumberOfFixedFurniture \\
PoliceOfficers & NumberOfPoliceOfficers & PublicToilets & NumberOfPublicToilets \\
CrimeRate & PerceptionOfCrimeRate & TransportationStations & NumberOfTransportationStations \\
CrimeSecurityDuringDay & PerceivedDaytimeCrimeSecurity & Greenness & ProportionOfGreenness \\
CrimeSecurityAtNight & PerceivedCrimeSecurityAtNight & WalkableSpace & ProportionOfWalkableSpace \\
Lights & NumberOfLights & DiverseLandscape & LandscapeDiversityIndex  \\
Graffiti & GraffitiOnBuildings & Colorfulness & EnvironmentalColorDiversity \\
FootTraffic & PerceptionOfPedestrianFlow & Sky & ProportionOfSky \\
LandmarkVisibility & LandmarkVisibilityIndex & Cleanliness & StreetCleanliness \\
DiverseLandscape & LandscapeDiversityIndex & LandmarkVisibility & LandmarkVisibilityIndex \\
Colorfulness & EnvironmentalColorDiversity & Transparency & TransparencyIndex \\

\bottomrule
\end{tabular}%
} }
\end{table*}
\end{tiny}

\section{Methodology}\label{sec:method}

This section introduces our methodology for MLLM-based street environment walkability evaluation.
Our methodology was shaped by the following scopes and aims: 
\paragraph{Aims} In the paper, our goal is to define expert knowledge in walkability assessment in distinct levels based on current literature, and to assess and compare the evaluative results provided by the MLLMs. This paper does not aim to propose a new set of indicators or refine the semantic correctness of indicator names from the literature; these aspects will be addressed in future work.

\paragraph{Scopes} We collect metrics that are directly related to public spaces, along with the criteria used to cap them, as documented in the existing literature. We conducted a comparative study designing four sets of questions. The prompt sets have varying levels of semantic clarity to represent different levels of embedding of expert knowledge. We applied inferential statistics to identify differences in MLLM assessments and highlight metrics with statistically significant variations between models, hence identifying the influence of intergating expert knowledge to MLLMs.

\subsection{Selecting and structuring walkability metrics}
\label{metrics}

We selected our walkability metrics and identified the criteria based on two review articles about measuring walkability \citep{dragovic2023LRofparameter-basedmodelforWB,fonseca2022builtenvironmentattributes}. Based on the combined literature sets from both reviews, we conduct a review of studies that develop walkability metrics. This combined set was used to develop a list of indexes, aimed at building upon existing work and reducing bias in the indicator selection process. Our work reveals that varying interpretations of what defines a good walking environment have led to a broad array of interchangeably used terms. 
For example, inconsistencies were observed where different metric names referred to similar measurements, or conversely, identical metric names were used despite differences in the underlying methods.

From our initial literature set, we collected 124 walkability metrics. 
We use \textit{Metric} to refer to direct measurables of the street environment, \textit{Methods} to refer to evaluation methods (e.g. a survey or a tool), \textit{Criterion} to refer to specific criteria in the evaluation approaches (e.g. Accessibility, Attractiveness). We introduced \textit{DataSource} class and \textit{ScoringFunction} class. 
We applied the Triple-A ontology \citep{pieter2021TripleA} to hierarchically structure and describe the final versions of these metrics, similar to the metric structuring described by \cite{grisiute2024ontology} and \cite{atamanmulti2022Multi-Criteria}.
We set out the categorisation as the first step towards structuring walkability evaluation metrics. The entire metric database is available \citep{Cai2025dataset}.

In this paper, we take metrics that are used for evaluating \textit{Safety} and \textit{Attractiveness} as examples and feed multi-modal large language models (MLLM) to make initial tests on their performance in evaluating the street environment walkability through SVIs. 
As shown in Table \ref{tab1}, we identified 21 metrics for each \textit{Criterion}. 
We then adapted these into two sets of metrics: one with vague definitions while naming the metrics and another with more quantifiers in the metric names. 
These metric lists, developed based on the current literature, are non-exhaustive for evaluating walkability, safety, and attractiveness but are used as comparison sets to test MLLMs' performance in evaluating the street environment.

\subsection{Shaping prompts for SVI samples}

\begin{figure}[ht]
\centering
\includegraphics[width=\columnwidth]{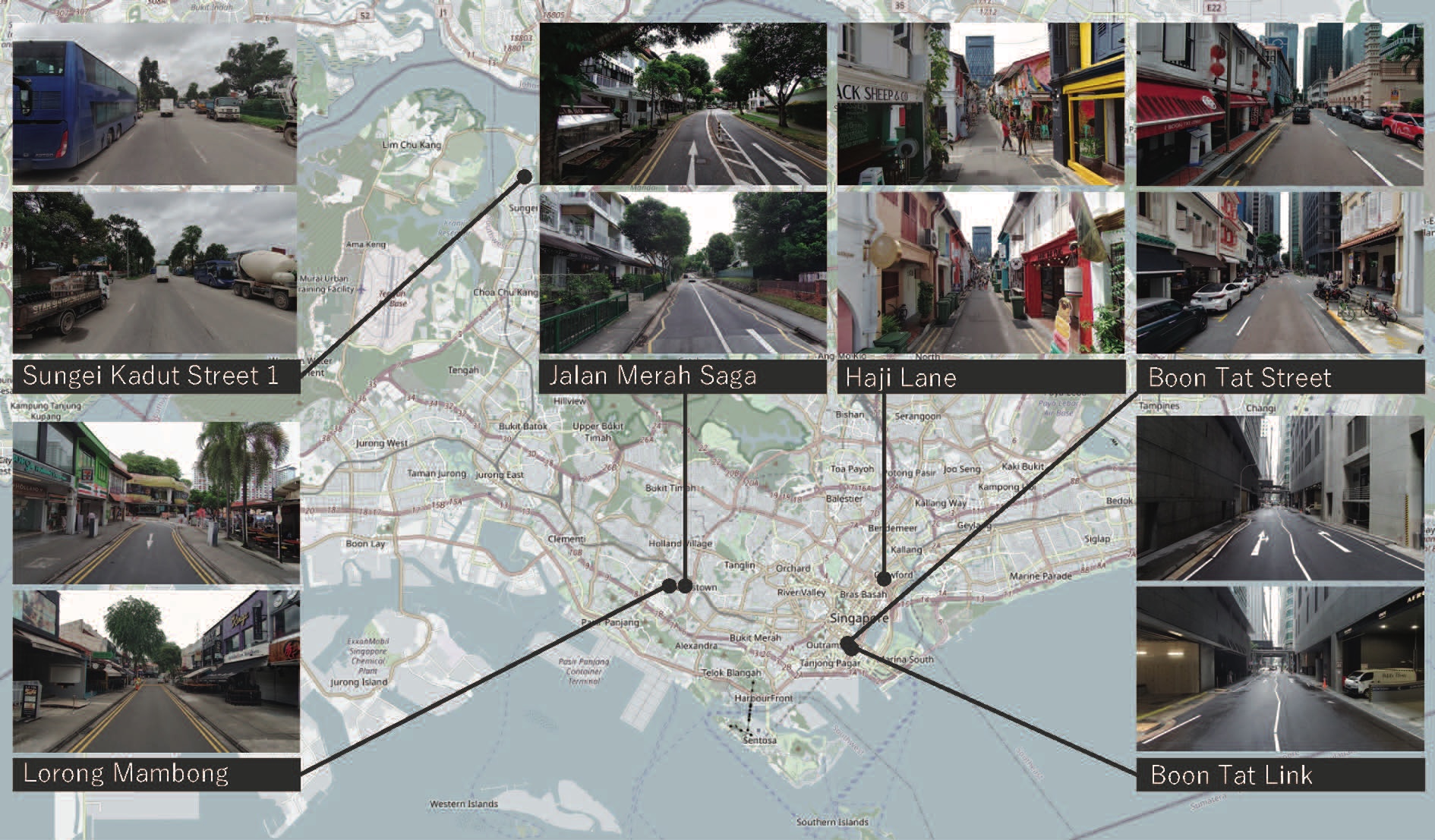}
\caption{Streets in Singapore were selected for evaluating SVI walkability, including streets from the downtown area, countryside, commercial centre, and residential areas. (c) KartaView contributors} \label{locations}
\end{figure}

We selected streets representing various types in Singapore, including locations from the downtown area, countryside, commercial centre, and housing areas (Figure \ref{locations}). A total of 42 SVIs are sourced from a crowd-sourced platform KartaView. 

The ChatGPT-4 model was used to evaluate the image dataset. Drawing on the metrics outlined in Section \ref{metrics}, we developed four distinct prompts, with different levels of defined information (which is one way of representing different levels of expertise), to assess the influence of formal definitions of expert knowledge on the walkability assessment using MLLM.
Correspondingly, we designed four models using prompts with four levels of semantic clarity and embedding of expert knowledge.  The models demonstrate an increasing expert level of evaluation metric definition and semantic clarity from Model-C1 to Model-C4. 
Model C1 has the lowest expertise level (level 1) and Model C4 has the highest expertise level (level 4).
Model-C1 simply asks GPT-4 to assess Safety and Attractiveness without any metric input.
Model-C4 prompt has a document of formal descriptions for each metric with more clarified definitions and scoring models. Table \ref{metricsDes} shows three examples of metrics and their descriptions. The full document is available \citep{Cai2025dataset}.

\begin{itemize}
    \item Level 1 of expertise : Model-C1 uses no metrics, asking GPT-4 to rate pedestrian \textit{Safety} and \textit{Attractiveness} on a scale from 1 (lowest) to 105 (highest) without specifying any evaluation metrics.
    \item Level 2 of expertise: Model-C2 uses the metrics from literature, but in vague language, incorporating Vague Safety Metric-1 and Vague Attractiveness Metric-1 with ratings on a scale from 1 (lowest) to 5 (highest) for each metric.
    \item Level 3 of expertise: Model-C3 uses metrics with quantifiers, incorporating Quantified Safety Metric-2 and Quantified Attractiveness Metric-2 with ratings on a scale from 1 (lowest) to 5 (highest) for each metric.
    \item Level 4 of expertise: Model-C4 uses quantified metrics and formal descriptions, integrating Quantified Safety Metric-2 and Quantified Attractiveness Metric-2, along with a document containing specified descriptions for each metric.
\end{itemize}

\begin{table}[ht]
    \caption{Three examples Model-C4 prompt, including metrics and their descriptions. The descriptions outline the definitions and scoring methods. }
    \label{metricsDes}
    \centering
    
    \begin{tabular}{p{\columnwidth}}
    \toprule
    \textbf{Examples} \\
    \midrule
    \textbf{Metric}: PresenceOfInstitutionalArea  \\
    \textbf{Description}: Institutional area refers to educational, medical, community and cultural areas. \\
    \textbf{Scoring}: If one of the above institutional areas is present, score: 5. If not, score: 1. 
    \\
    \midrule
    {\textbf{Metric}:NumberOfTrafficCalmingDevices} \newline
    \textbf{Description}:The number of traffic calming devices, such as speed bumps, raised crosswalks, and pedestrian islands. \newline
    \textbf{Scoring}:A higher number of traffic calming devices corresponds to a higher score. \\
    \midrule
   {\textbf{Metric}: GraffitiOnBuildings} \newline
    \textbf{Description}: If graffiti is present on buildings, indicating a sense of unsafety, score: 1.\\
    \bottomrule
    \end{tabular}
    \label{tab:my_label}
\end{table}

The models are tested in order from Model-C1 to Model-C4, with metrics and descriptions introduced only in the final model to prevent the language model from learning from the descriptions and influencing the results of other models. 
All tests are conducted on the same machine by the same user to avoid discrepancies between machines and accounts and to ensure the comparative study's consistency and reliability. 

To further evaluate the scoring behaviour of the four language models under different prompts, Levene's Test is conducted to assess the assumption of homogeneity of variances as a prerequisite for variance analysis. Subsequently, Welch's ANOVA and the Games-Howell post hoc test are employed to examine differences in score distributions, both for the overall score and across individual metrics. 
Among the 21 metrics analyzed, the six exhibiting the most significant statistical differences are identified and further investigated to gain insights into the models' interpretative behaviour. 
Based on the statistical results, we further investigate the case studies of places with lower scores and discuss the potential urban design interventions.

\section{Results and Discussion}

\subsection{Results of MLLM evaluations on SVIs}

\begin{figure}[ht]
\centering
\includegraphics[width=0.55\columnwidth]{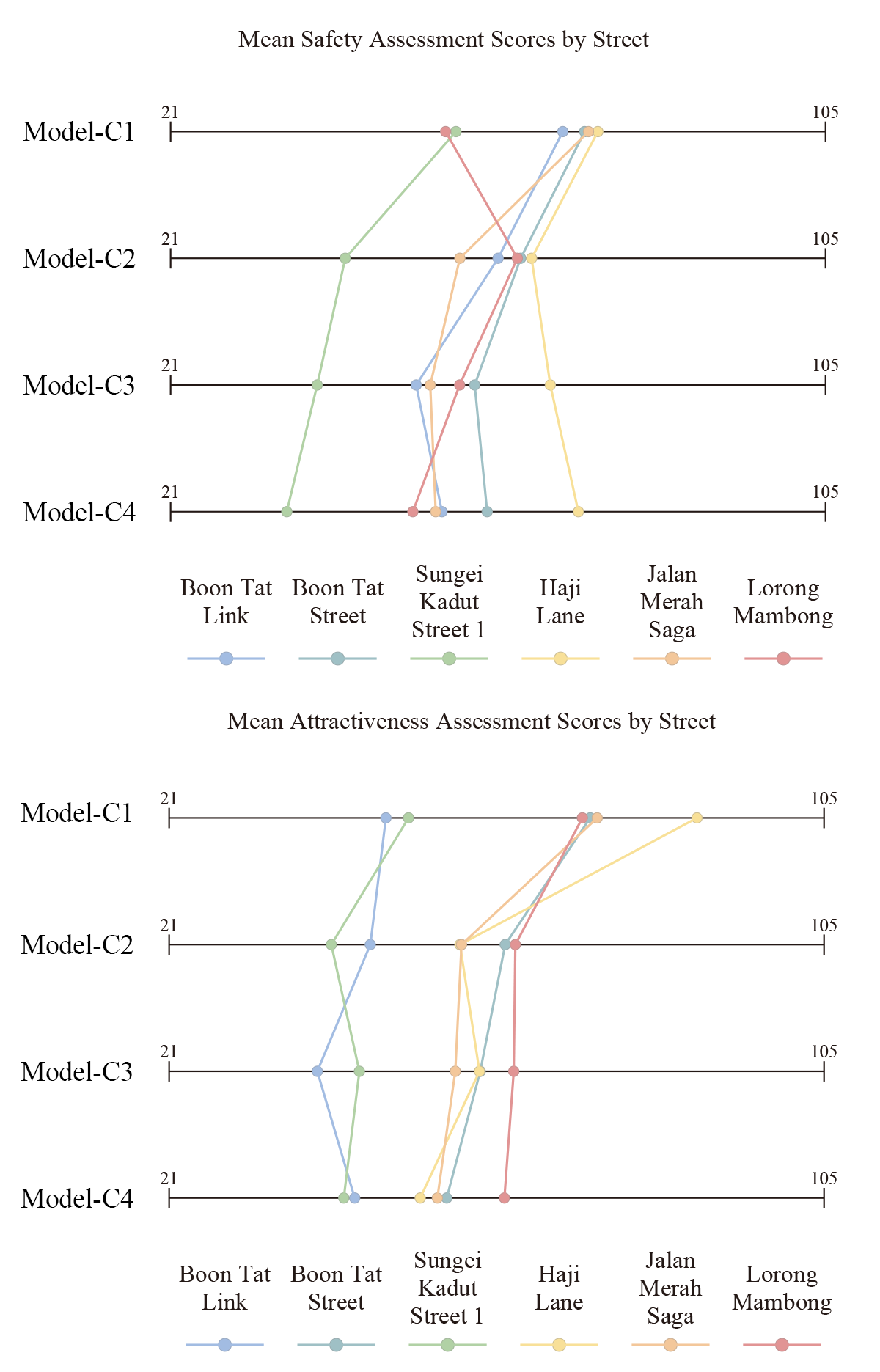}
\caption{Plot of the streets' average scores in safety and attractiveness assessments, as evaluated by the four MLLMs. 
} \label{StreetAve}
\end{figure}

Figure \ref{StreetAve} shows the average safety and attractiveness scores for each street across the four models. Model-C1 diverges significantly from the other models in both assessments. It shows notable differences in scoring ranges across the six streets, too. For instance, Lorong Mambong scores lowest in safety in Model-C1 but ranks mid-range in other models.

\begin{figure}[ht!]
\centering
\includegraphics[width=0.6\columnwidth]{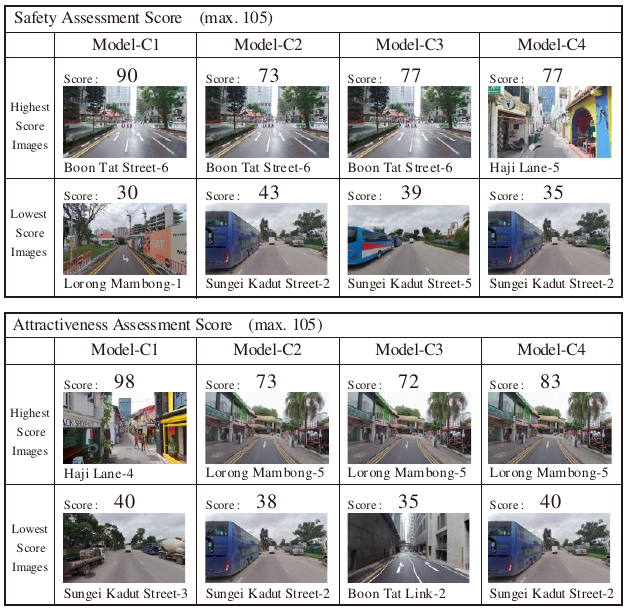}
\caption{Street view images with the highest and lowest scores for safety (top) and attractiveness (bottom), giving the corresponding scores. } \label{highestlowest}
\end{figure}

Figure \ref{highestlowest} presents the SVIs that received the highest and lowest scores across the four models. Overall, the models similarly depict the safest and most attractive streets, as well as the least safe and least attractive streets.
In the safety assessment, Haji Lane-5 achieved the highest score in Model-C4, while Boon Tat Street-6, which ranked highest in the other models, received the second-highest score (75) in Model-C4, showing the similarity between the four models.
These results provide an initial indication that while MLLM can offer assessments based on general knowledge, integrating MLLM with representations of expert knowledge significantly influences its evaluative performance.

To compare the performance differences between the models, we conducted statistical analysis on all the SVI assessment scores generated from all four models.
We conducted Levene's Test for the four models applied for \textit{Safety} and \textit{Attractiveness} to assess the equality of variances. In both cases, the null hypothesis was rejected.
Consequently, we employed Welch's ANOVA to evaluate the differences in distributions across the four models for each metric. The results revealed statistically significant differences, with p-values for both tests below 0.01. We then performed the Games-Howell post hoc test and visualised the score distributions within 95\% confidence intervals.

\begin{footnotesize}
\begin{table}[ht]
\centering
\caption{Statistical comparison of MLLMs' scoring distributions for safety and attractiveness using Welch ANOVA. p-val(S) is the p-value from safety scores, and p-val(A) is the p-value of attractiveness scores.}
\label{anova}
\begin{tabular}{p{3.0cm}p{3.0cm}p{2cm}p{2 cm}}
\toprule
\multicolumn{2}{c}{\bfseries Inter-group Comparisons} & \bfseries p-val(S) & \bfseries p-val(A) \\
\midrule
Model-C1 & Model-C2 & $<$0.01*** & $<$0.01*** \\
Model-C1 & Model-C3 & $<$0.01*** & $<$0.01*** \\
Model-C1 & Model-C4 & $<$0.01*** & $<$0.01*** \\
Model-C2 & Model-C3 & 0.20 & 0.99 \\
Model-C2 & Model-C4 & 0.27 & 0.51 \\
Model-C3 & Model-C4 & 0.99 & 0.57 \\
\midrule
\multicolumn{2}{c}{\bfseries Overall Test Statistics} & $<$0.01*** & $<$0.01*** \\
\bottomrule
\end{tabular}
\end{table}
\end{footnotesize}

\begin{figure}[ht]
\centering
\includegraphics[width=0.55\columnwidth]{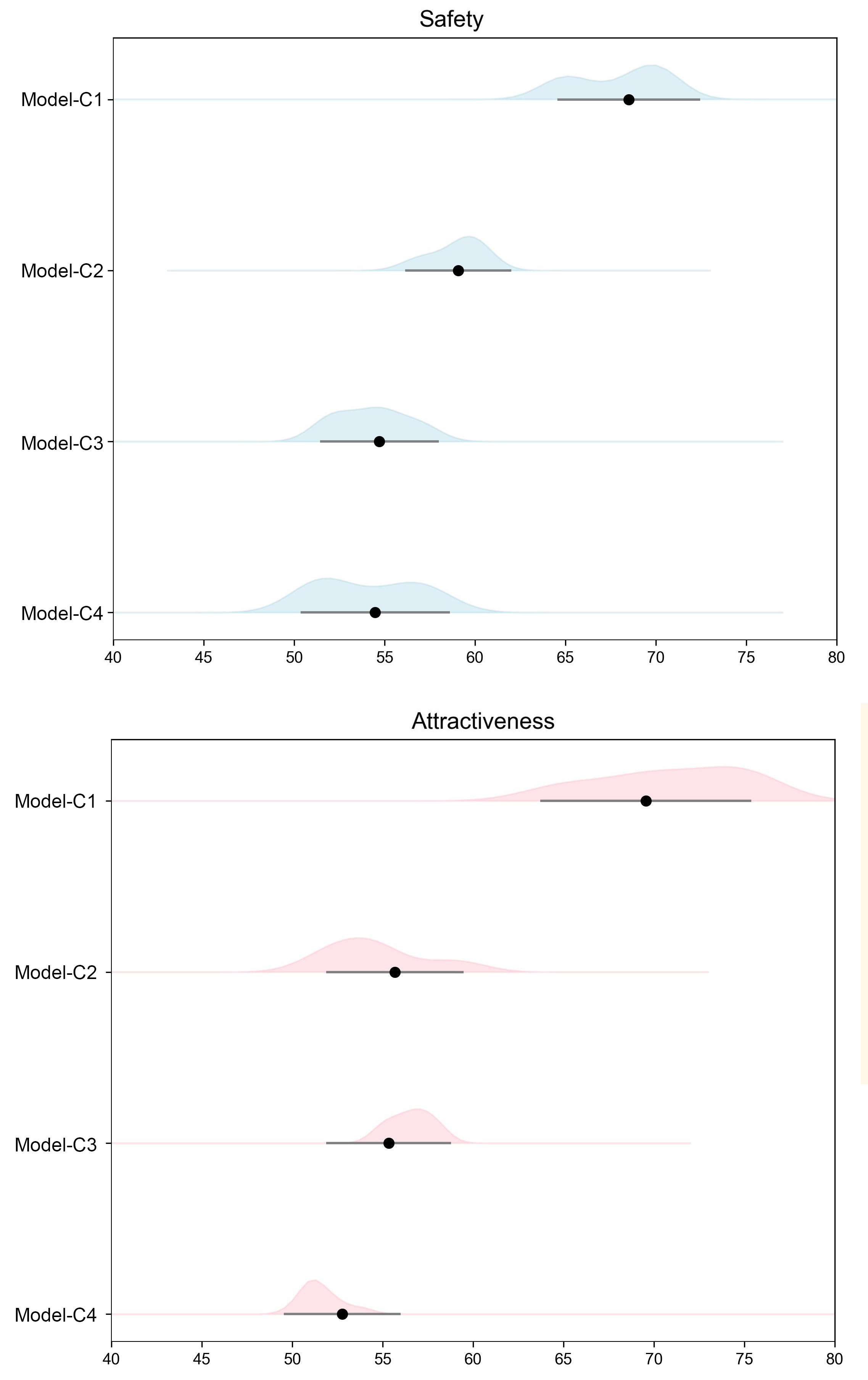}
\caption{The score distributions of the four MLLMs assessing safety (top) and attractiveness (bottom).  } \label{distribution}
\end{figure}

From the score distributions (Figure \ref{distribution}) and post hoc test results (Table \ref{anova}), Model-C1 produced significantly higher scores, suggesting that in the absence of detailed instructions, the language model tends to produce more optimistic assessments. 
In contrast, the other three models show no significant differences in their distributions, while their median scores exhibited slight variations in different directions. 
This indicates that adding evaluation metrics with specific criteria can significantly influence MLLM's assessments. However, increasing the level of semantic clarity by adding descriptions, has a relatively less pronounced impact on the overall scores.

\begin{figure*}[ht]
\centering
\includegraphics[width=\textwidth]{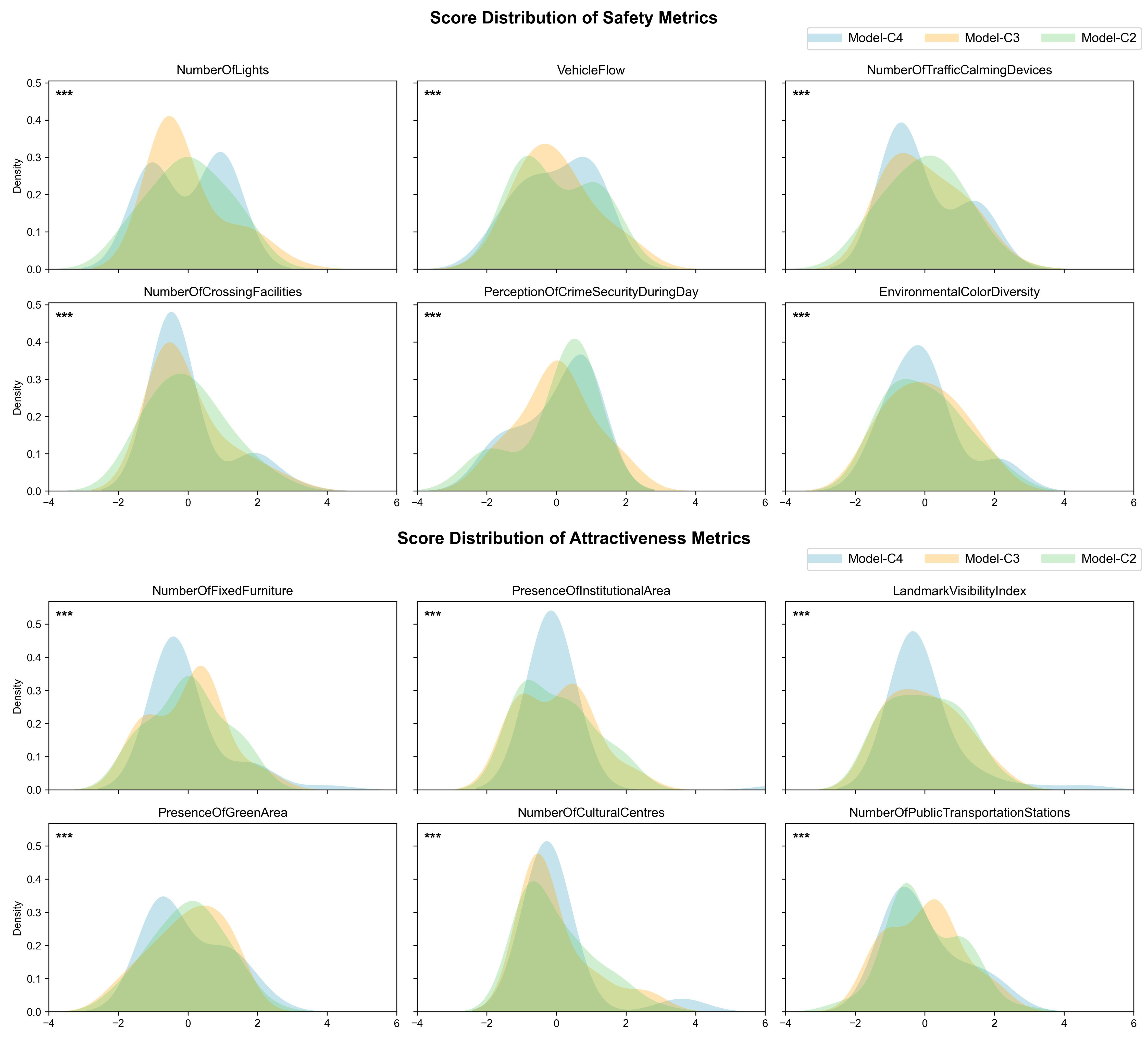}
\caption{The top six metrics with the largest statistical differences according to safety scores (top) and attractiveness scores (bottom), ranked by their test statistics. The density plots show the level of concentration in each MLLM when measuring the particular metric.} \label{metrics}
\end{figure*}

To further investigate the difference between the models with input metrics, we analyzed the scoring differences for each metric assessed by Model-C2 (with vague metrics), Model-C3 (with quantified metrics), and Model-C4 (with quantified metrics and descriptions).
We conducted the Kruskal-Wallis Test for the three models. 
Figure \ref{metrics} shows the top six metrics with the largest statistical differences under the \textit{Safety} and \textit{Attractiveness} assessments, ranked by their test statistics. In most metrics shown in Figure \ref{metrics}, the score distribution for Model-C4 exhibited a higher degree of concentration. 
This trend might be attributed to the more defined descriptions provided along the metrics in Model-C4, which likely enhance the MLLM's ability to analyze according to the specific definition of each metric. 

\begin{figure}[ht]
\centering
\includegraphics[width=0.7\columnwidth]{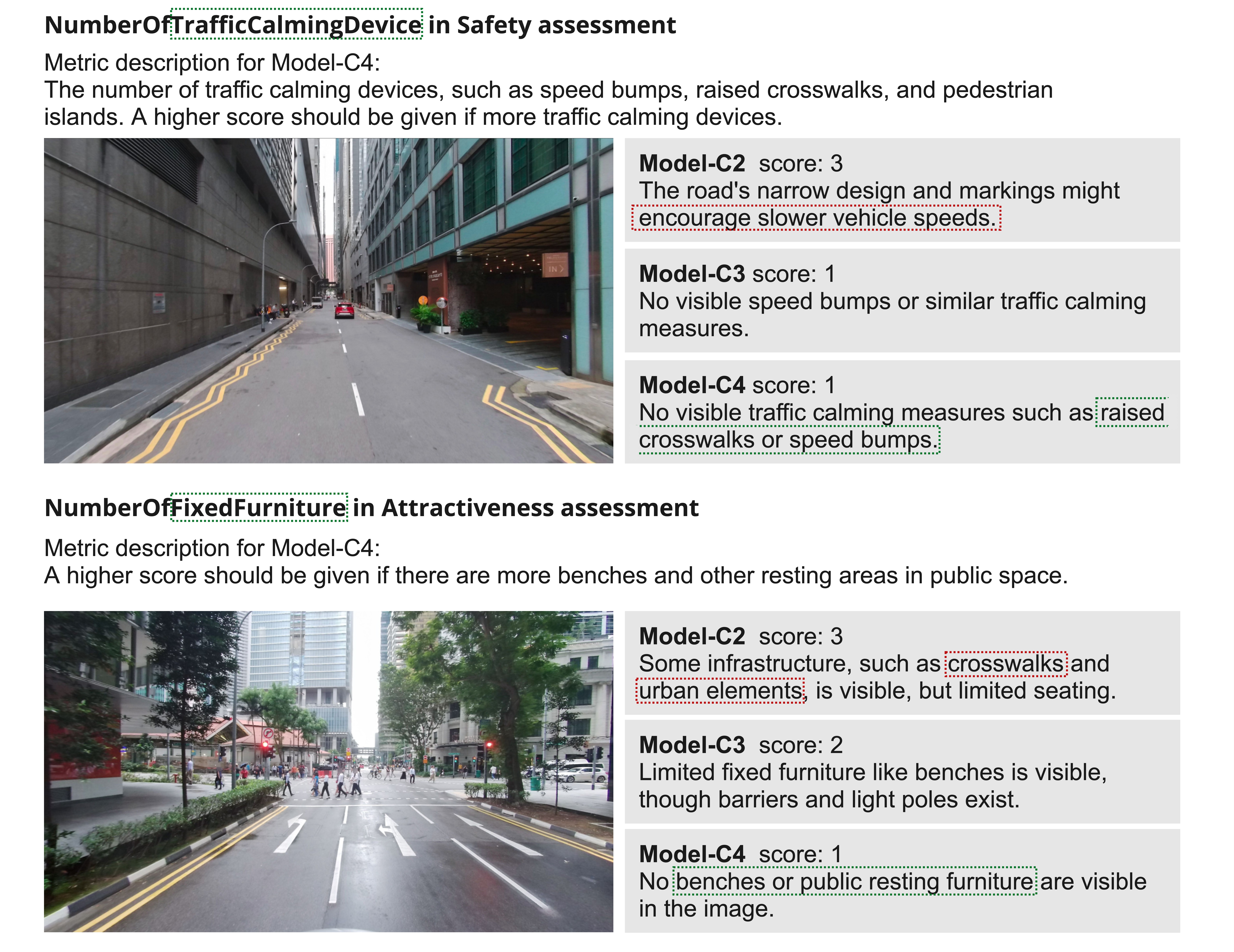}
\caption{Two example responses from MLLMs for the metrics that differ significantly among the three models, assessing safety (top) and attractiveness (bottom). This shows the influence of descriptions to MLLM evaluations.} \label{showcase}
\end{figure}

\newpage
To further investigate the influence of descriptions on MLLM evaluations, we examined two examples shown in Figure \ref{showcase}.
In the \textit{Safety} assessment, Model-C4 assigned a low score to the image because the specified traffic calming devices (e.g., speed bumps and raised crosswalks) were not visible. In contrast, Model-C2 inferred the presence of traffic calming devices based on narrow roads and markings, resulting in a more positive score.
Similarly, in the attractiveness assessment, Model-C4 assigned scores based on the descriptions of \textit{FixedFurniture}, while Model-C2 mistakenly considered crosswalks as a type of fixed furniture, leading to an incorrect evaluation.

Therefore, the more varied and dispersed score distributions in Model-C3, which uses only quantified metrics, and Model-C2, which uses vague metrics, could be attributed to the ambiguity resulting from the lack of definitions, leaving room for MLLM's interpretations.
In contrast, the detailed descriptions in Model-C4 reduce ambiguity in the prompts, thereby guiding the model's interpretations with the intended evaluation criteria, resulting in higher concentration and consistency.

\subsection{Implications of urban design intervention}

\begin{figure*}[ht!]
\centering
\includegraphics[width=0.8\textwidth]{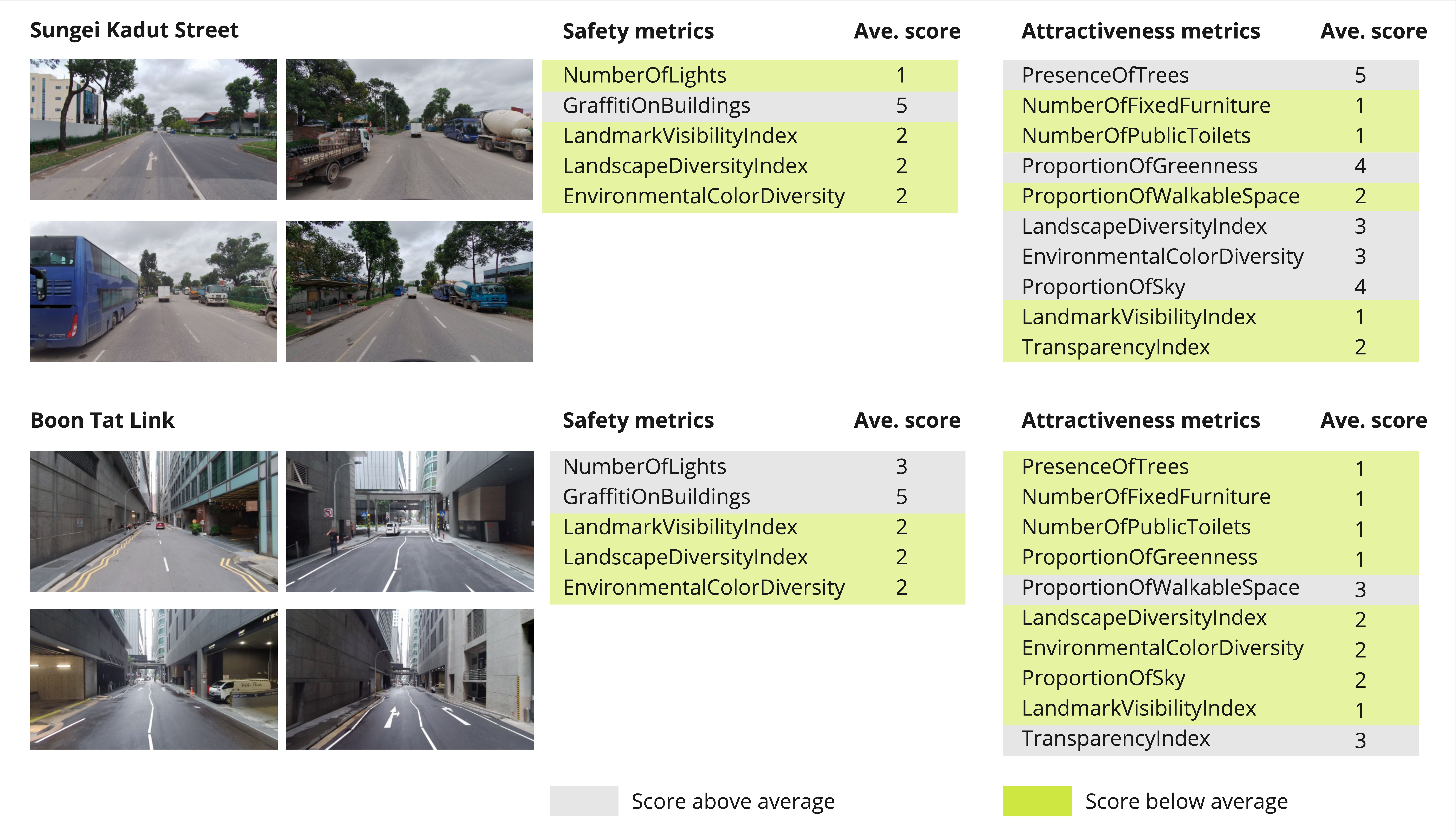}
\caption{Identifying urban design actionable metrics within the safety and attractiveness evaluations. The figure highlights the lower scores in the metrics, providing evidence to inform design interventions.} \label{udmetric}
\end{figure*}

To enhance the impact of assessments on urban design, it is essential to categorise evaluation metrics based on their applicability to urban design actions, thereby identifying strategies for improvement. 
Many metrics used in assessing \textit{Safety}, such as the PresenceOfPedestrianSignals and NumberOfTrafficCalmingDevices, fall outside the scope of urban design and are instead determined by other urban governance sectors, such as transportation authorities. 
The assessment of \textit{Attractiveness} includes slightly more metrics actionable through urban design, such as PresenceOfTrees, NumberOfFixedFurniture, and EnvironmentalColorDiversity. 
Metrics that are not actionable through urban design have been filtered out, as illustrated in Figure \ref{udmetric}.

Based on the previous assessments of the six streets, Sungei Kadut Street (StreetS) and Boon Tat Link (StreetB) are identified as having lower scores for safety and attractiveness. By focusing on the urban design actionable metrics associated with their lower scores, targeted interventions can be proposed to improve these streets (Figure \ref{udmetric}). 
For instance, both streets score low in safety concerning factors such as lighting, landmark visibility, and environmental colour diversity. However, StreetB performs slightly better due to the presence of lighting.
In the attractiveness assessment for StreetB, its low score can be attributed to the lack of trees, resting areas, landmarks, and the limited sky visibility caused by the narrow spacing between buildings.
Improvements could focus on adding trees, incorporating landmarks, or enhancing the street's visual appeal through the use of colour. 
Generally, urban design indicators exhibit greater levels of variation in different semantic clarity levels, compared to non-urban design indicators. 
In the metric difference analysis, if we exclude presence-based indicators (as their binary scoring models can cause larger discrepancies), a higher proportion of urban design indicators rank among the top.

\subsection{Discussion}
We discuss the findings of the paper in three aspects. First, the results offer preliminary evidence that MLLMs can perform assessments based on the provide evaluation metrics, using general knowledge. The multimodal model can automatically provide interpretations and scoring based on images. For example, it assumed the perceived traffic flow based on the width of the road. Therefore, MLLMs are potentially helpful for large-scale studies of urban design quality assessment. 
Second, Model-C1 produces more optimistic scores compared to the other three models. The results of its evaluation for the street segments show discrepancies compared to the others. For example, a street that receives the lowest safety score in Model C1 is ranked mid-range in the other models. This suggests that MLLM evaluations without incorporating expert knowledge tend to be overly optimistic and may produce feedback that diverges from expert-informed evaluations.
Third, informing MLLMs with expert knowledge (e.g., evaluation metrics) significantly influences their evaluative performance. For example, enhancing the semantic clarity of the walkability metrics described in the prompts helps align the MLLM interpretations with the specified criteria, leading to increased concentration and consistency. 
Although the addition of descriptions to the metrics has a comparatively less pronounced effect on overall scores, it can prevent MLLM from making mistakes when interpreting the provided metrics. 

The paper presents our initial exploration of the risks and advantages of large language models in evaluating environmental quality and the potential to integrate expertise and domain knowledge. 
MLLMs offer a low entry barrier for generating automated multimodal image-text assessments of walkability and can support large-scale evaluations of the quality of urban design. While this automation holds promise for broad, data-intensive studies, it also presents risks, such as overly optimistic evaluations, misinterpretation of terminology, and inaccurate assessments. Therefore, further research is needed to better understand how MLLM can be effectively integrated into urban design quality surveys and to establish standardised guidelines for prompt formulation.

\section{Conclusion}

This study explores how integrating more formal and clarified representations of expert urban design knowledge into the input prompts of an MLLM (ChatGPT) can enhance its capability to evaluate walkability using SVIs. 
Walkability metrics were collected and categorised through the existing literature. 
A comparative study was conducted for MLLMs fed with prompts with varying levels of clarity and specificity, using the metrics for assessing pedestrian safety and attractiveness.
The findings demonstrate that MLLMs' evaluative performance can be enhanced by integrating expert knowledge. Furthermore, increasing the semantic clarity of expert knowledge representations improves the consistency of MLLMs' evaluative outputs. 

This paper has limitations of the small size of the SVI database and the limited set of metrics for walkability evaluation. These challenges mark the starting points for future enhancements and developments in this area. Our future work will be extended from three aspects: 1) engaging urban design practitioners to evaluate the SVIs and compare their assessment with the MLLMs using expert knowledge from literature; 2) establishing a more rigorously defining walkability criteria through the lens of urban design assessment characteristics and developing a walkability assessment ontology; 3) scaling up the database to support an automated evaluation workflow that can provide urban designers with practical guidelines.

\paragraph{Acknowledgement}
Part of this research was conducted at the Future Cities Lab Global at Singapore-ETH Centre. Future Cities Lab Global is supported and funded by the National Research Foundation, Prime Minister’s Office, Singapore under its Campus for Research Excellence and Technological Enterprise (CREATE) programme and ETH Zurich (ETHZ). 

%
%

\bibliographystyle{splncs04}
\bibliography{reference}

\end{document}